\documentclass{article}
\usepackage{spconf,amsmath,graphicx}

\newcommand{\eg}{{\emph{e.g.}}}

\newlength\savedwidth

\makeatletter
\newcommand*{\rom}[1]{\expandafter\@slowromancap\romannumeral #1@}
\makeatother

\usepackage{amsfonts}
\usepackage{amsmath}
\usepackage{array}
\usepackage{textcomp}
\usepackage{url}
\usepackage{verbatim}
\usepackage{color}
\usepackage{multirow}
\usepackage{setspace}
\usepackage{dsfont}
\usepackage{bm}
\usepackage{makecell}
\usepackage{graphicx}
\usepackage{caption}
\usepackage{subcaption}
\usepackage{booktabs}

\title{Context and Pixel Aware Large Language Model \\for Video Quality Assessment}
\name{Wen Wen$^{1\ast}$, Yaohong Wu$^{2}$, Yue Sheng$^{2}$, Neil Birkbeck$^{2}$, Balu Adsumilli$^{2}$, and Yilin Wang$^{2\dagger}$\thanks{$^\ast$Work done during internship at Google. $\dagger$Corresponding author: yilin@google.com.}}
\address{$^{1}$City University of Hong Kong \\
         $^{2}$Google Inc.}
         
\vspace{-12pt}

\begin{document}
%\ninept
%
\maketitle
\begin{abstract}
Video quality assessment (VQA) is a challenging research topic with broad applications. 
Traditional hand-crafted and discriminative learning-based VQA models mainly focus on pixel-level distortions and lack contextual understanding, while recent multimodal large language models (MLLMs) struggle with sensitivity to small distortions or handle quality scoring and description as separate tasks. 
To address these shortcomings, we introduce \textbf{CP-LLM}: a \textbf{C}ontext- and \textbf{P}ixel-aware \textbf{L}arge \textbf{L}anguage \textbf{M}odel. CP-LLM is a novel multimodal LLM architecture featuring dual vision encoders designed to independently analyze perceptual quality at both high-level (video context) and low-level (pixel distortion) granularity, along with a language decoder that subsequently reasons about the interplay between these aspects.
This design enables CP-LLM to simultaneously produce robust quality scores and interpretable quality descriptions, with enhanced sensitivity to pixel distortions (\eg,~compression artifacts). 
Experiment results demonstrate that CP-LLM achieves state-of-the-art cross-dataset performance on VQA benchmarks and superior robustness to pixel distortions.
\end{abstract}

\section{Introduction}

Video quality assessment (VQA) is a fundamental research area with broad applications in video compression, transcoding, transmission, playback, content search, and recommendation systems.  
Early no-reference VQA models focused on low-level distortions like blur and blocking artifacts.
% ~\cite{saad2014blind, mittal2015completely, korhonen2019two, korhonen2020blind}. 
However, with the emergence of user-generated content (UGC) videos, several subjective studies~\cite{wang2019youtube, wang2024youtube} have demonstrated that identical low-level artifacts can elicit significantly different perceptual quality ratings.
Consequently, effective UGC-VQA necessitates a comprehensive analysis encompassing both low-level pixel distortions and high-level semantic information.
Recognizing this imperative, an increasing number of recent methodologies have aimed to explicitly integrate high-level semantic insights into their model architectures.
For instance, by combining hand-crafted signals with CNN-derived features~\cite{tu2021rapique} or using specialized subnets for content and distortion~\cite{wang2021rich}.

\begin{figure*}[t]
    \centering
    \begin{subfigure}[]{0.245\textwidth}
        \includegraphics[width = \columnwidth]{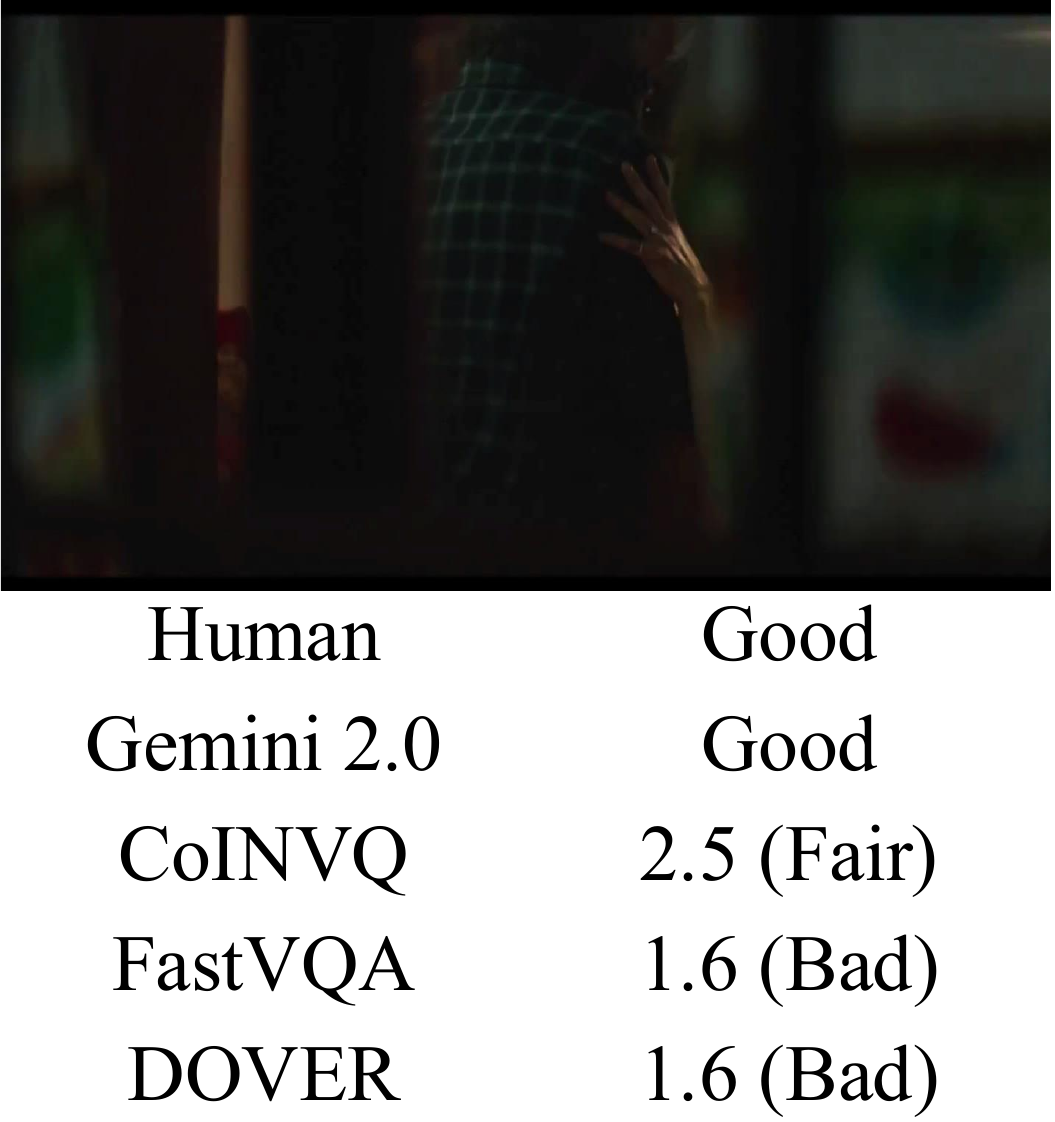}
        \label{fig:movie1}
    \end{subfigure}
    \hfill
    \centering
    \begin{subfigure}[]{0.245\textwidth}
        \includegraphics[width = \columnwidth]{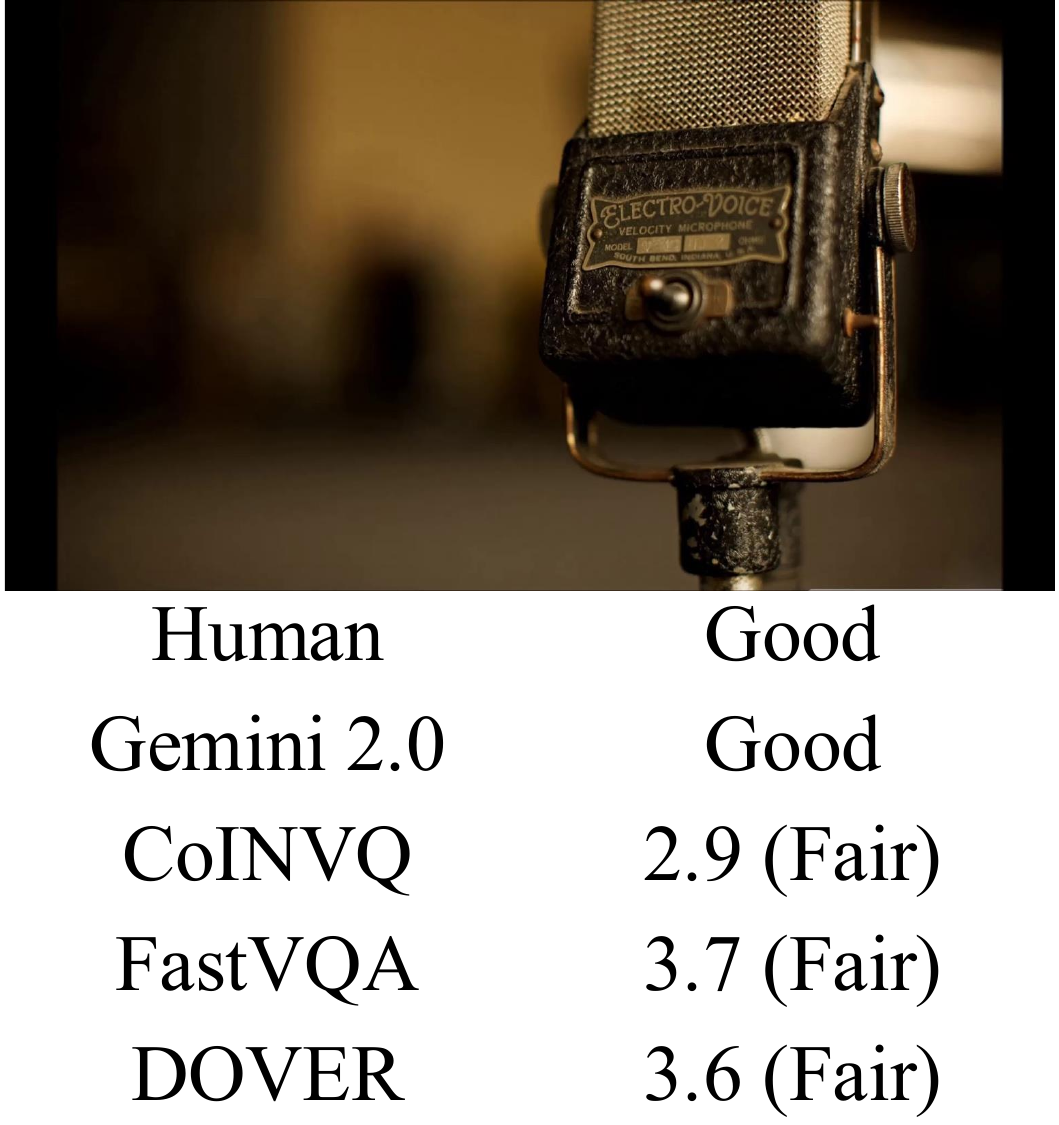}
        \label{fig:movie2}
    \end{subfigure}
    \hfill
    \centering
    \begin{subfigure}[]{0.245\textwidth}
        \includegraphics[width = \columnwidth]{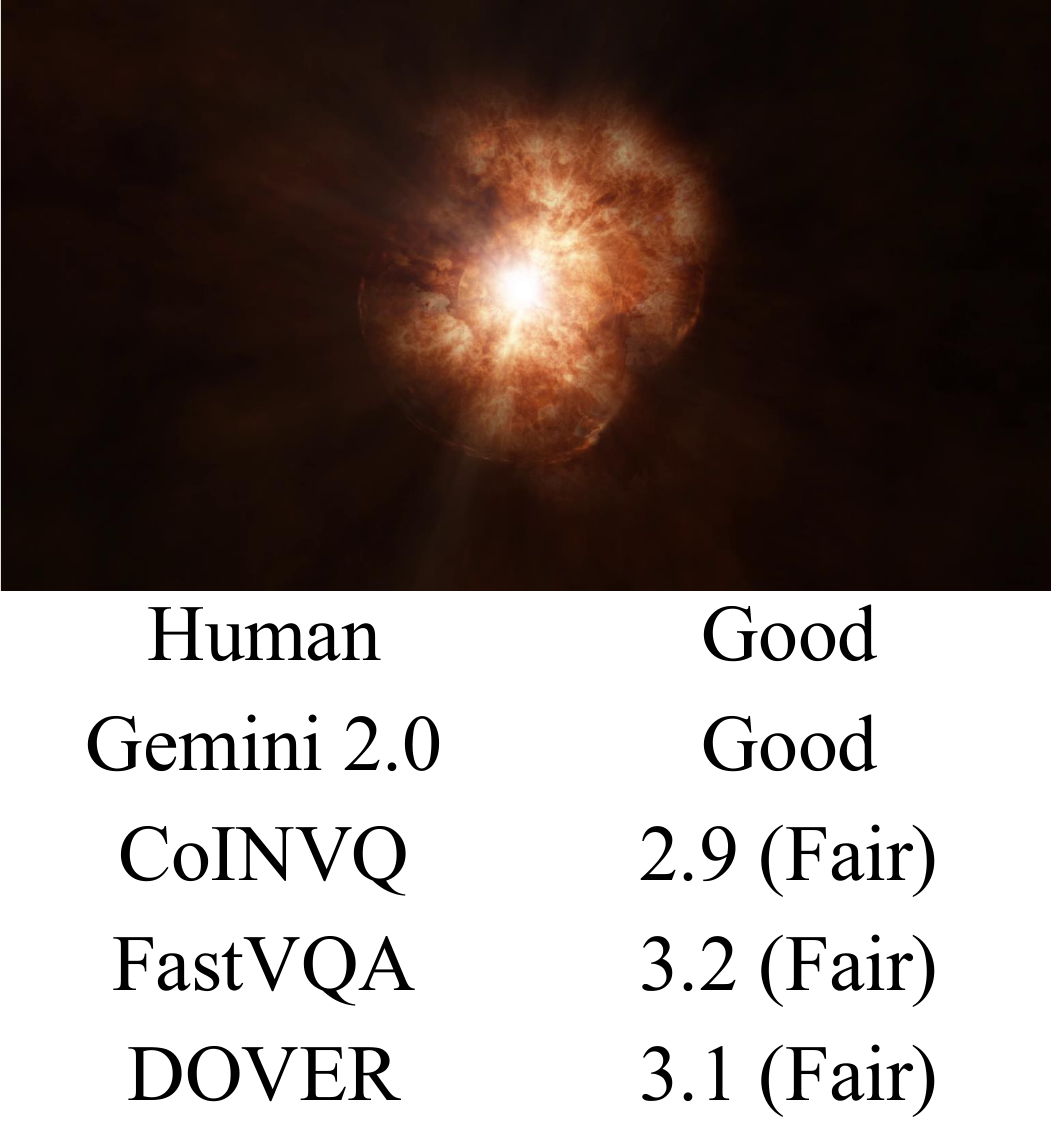}
        \label{fig:game1}
    \end{subfigure}
    \hfill
    \centering
    \begin{subfigure}[]{0.245\textwidth}
        \includegraphics[width = \columnwidth]{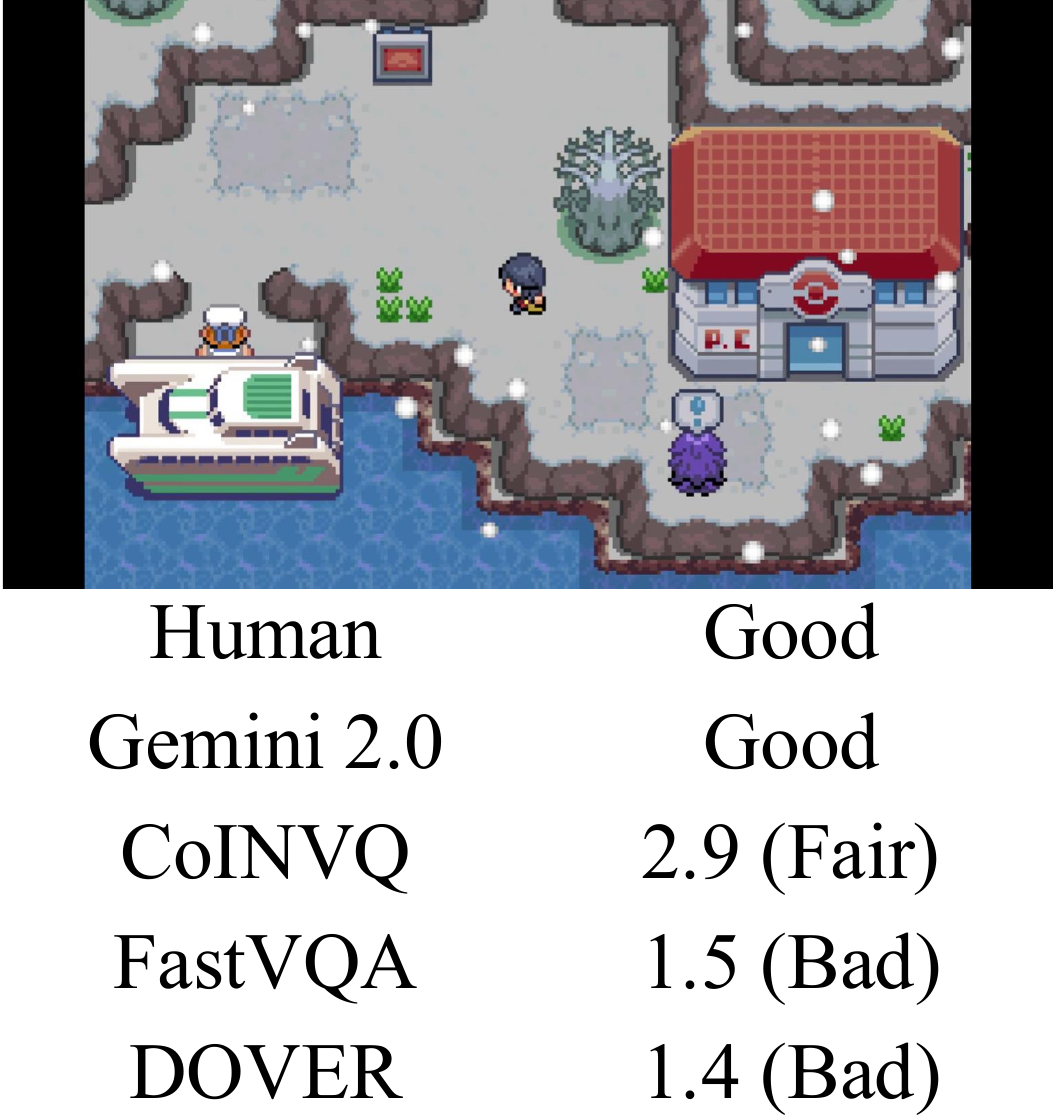}
        \label{fig:game2}
    \end{subfigure}
    \caption{Videos with intentional effects but incorrectly identified as quality defects by traditional VQA models (CoINVQ~\cite{wang2021rich}, FastVQA~\cite{wu2022fast}, and DOVER~\cite{wu2022disentangling}), while MLLM's assessment aligns better with human perception. The first two videos contain intentional low-light and background blurring.  The third video presents an astronomy video containing complex visual textures. The last video shows a pixel-art style game characterized by inherent blockiness. All quality scores are calibrated to $[1, 5]$, where $1$, $2$, $3$, $4$, $5$ correspond to bad, poor, fair, good, and excellent, respectively.  
    }
    \label{fig:movie_game}
    \vspace{-6pt}
\end{figure*}

Although the approaches mentioned above underscore the value of leveraging semantic signals, they possess limitations, notably in their content understanding capabilities. As illustrated in Fig~\ref{fig:movie_game}, certain visual artifacts (\eg, dark scenes, intentional blur, pixelation) that are integral to the artistic style or narrative of the content are often misinterpreted as quality defects by state-of-the-art (SOTA) VQA models such as CoINVQ~\cite{wang2021rich}, FastVQA~\cite{wu2022fast}, and DOVER~\cite{wu2022disentangling}. In contrast, advanced multimodal large language models (MLLMs) like Gemini and GPT have demonstrated the capacity to correctly interpret these stylistic choices as intended, thereby rating such examples as high quality and highlighting a crucial capability absent in many current VQA models.

\begin{figure*}[!t]
\centering
\includegraphics[width=0.85\textwidth]{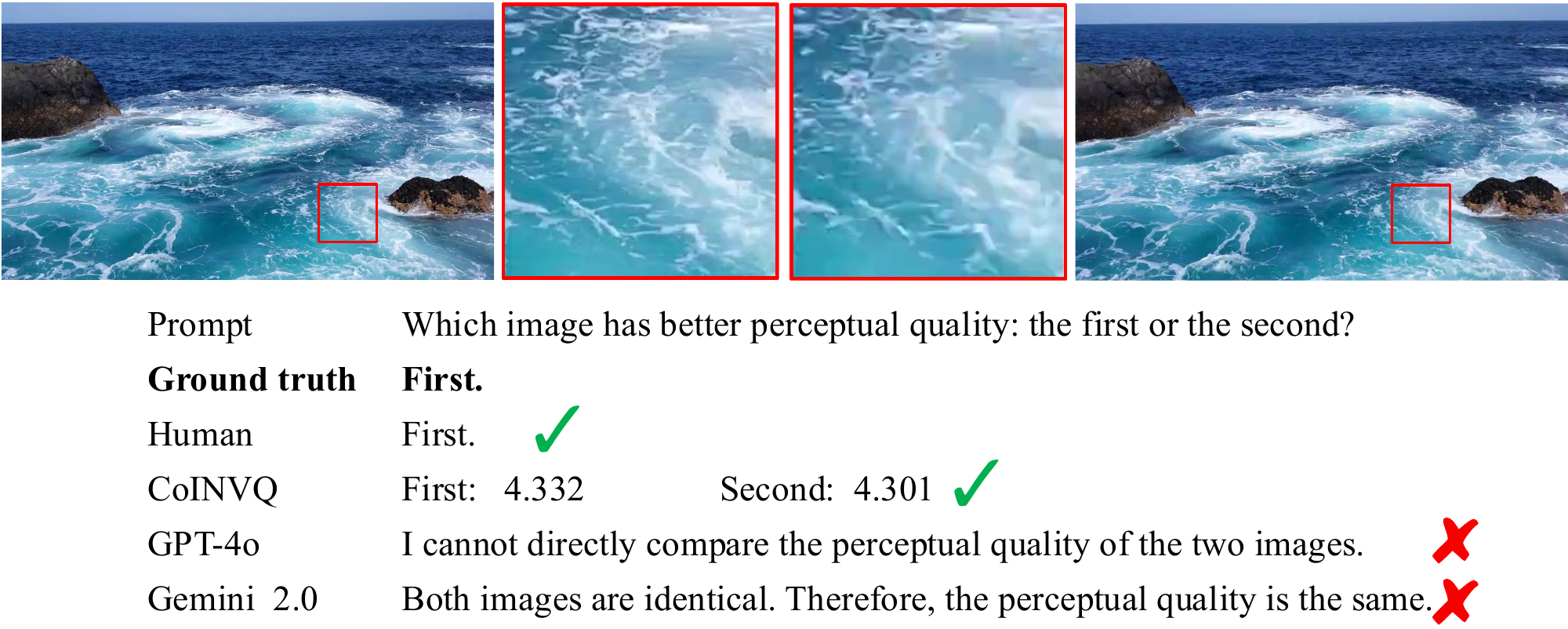}
\caption{
An original video (left) and its compressed version (right). The quality degradation is easily detected by human and VQA models (\eg~CoINVQ), but not correctly caught by MLLMs.}
\label{fig:orig_crf}
\vspace{-6pt}
\end{figure*}

The advent of MLLMs offers a promising new paradigm for VQA. Nevertheless, current MLLM-based approaches face significant challenges. Their vision encoders are often optimized for high-level semantic understanding rather than fine-grained perceptual analysis, which hinders the detection of subtle pixel-level distortions (see Fig~\ref{fig:orig_crf}). Furthermore, MLLMs typically process low-resolution inputs, requiring resizing operations that can introduce new artifacts by disregarding the original aspect ratio.

A majority of existing works employ MLLMs to perform either quality-related description generation or quality score prediction, but rarely both. 
While a few studies have used reinforcement learning to unify scoring and description, this exploration has been confined to image quality assessment, leaving a gap in the video domain.

To address these limitations, we introduce CP-LLM, a novel MLLM-based VQA framework designed to detect pixel-level artifacts while concurrently providing robust score predictions and relevant quality descriptions. Our key contributions are:
\begin{itemize}
    \item An MLLM-based VQA model capable of simultaneously generating quality score predictions and interpretable perceptual descriptions.
    \item A comprehensive experiment design that evaluates both accuracy and robustness of the VQA model, with novel evaluation criteria (\eg~flip rate) that have not been sufficiently addressed in existing VQA evaluations.
\end{itemize}

\begin{figure*}[!t]
\centering
\includegraphics[width=0.9\textwidth]{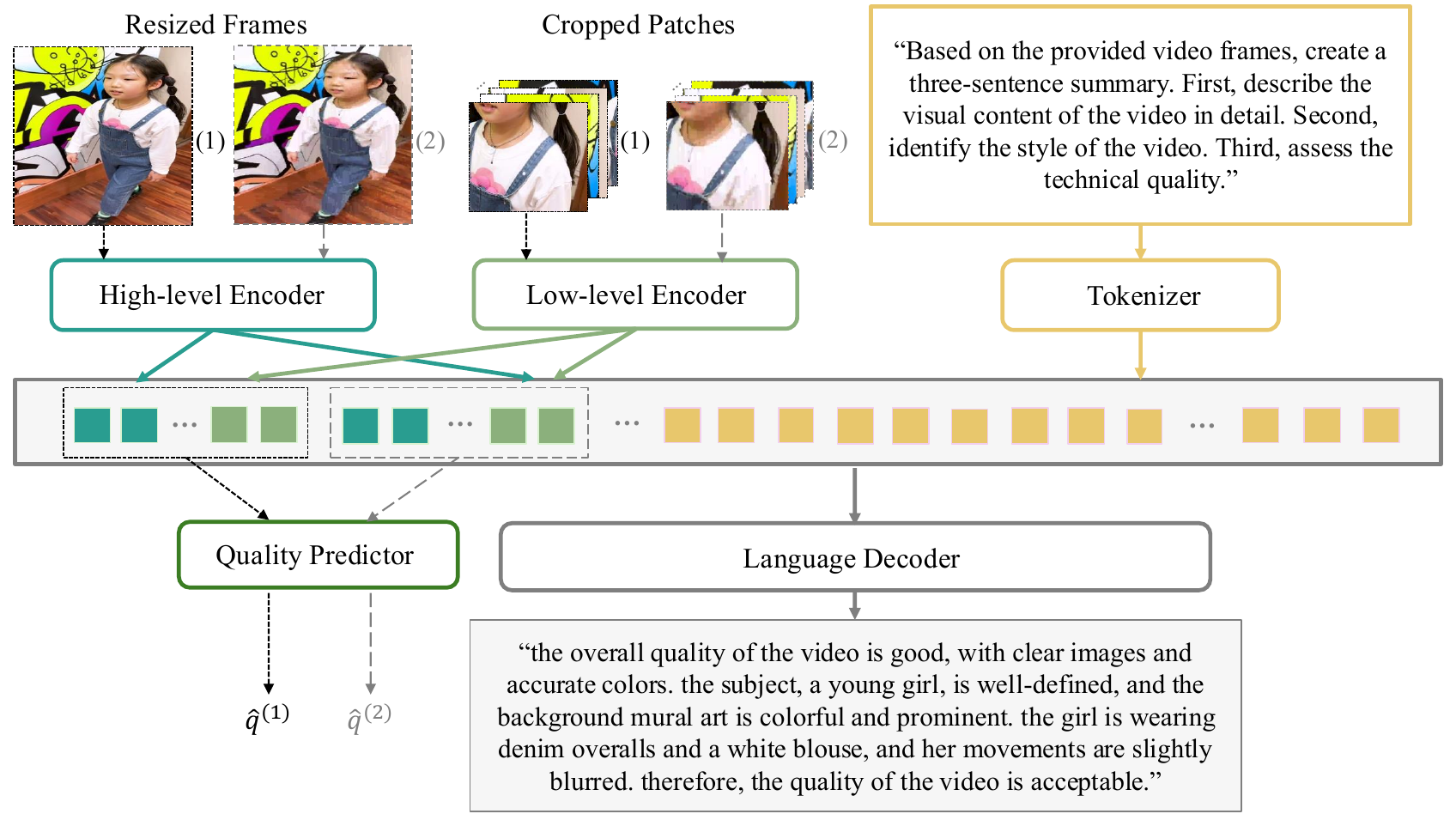}
\caption{The overview of CP-LLM  framework. The model processes video frames through two distinct visual pathways: a high-level encoder extracts semantic features, and a low-level encoder captures fine-grained perceptual details from image patches. Embeddings from both visual streams are interleaved with text prompt embeddings and fed into a language decoder to generate textual output. Concurrently, both the concatenated visual embeddings are processed by an MLP regression head to predict a numerical quality score. }
\label{fig:model}
\vspace{-12pt}
\end{figure*}

\section{Related Work}

\noindent\textbf{Knowledge-driven and Learning-based Methods.} Objective methods for assessing UGC video quality have evolved significantly in past decades. 
Early knowledge-driven approaches, such as TLVQM~\cite{korhonen2019two}, and VIDEVAL~\cite{tu2021ugc}, primarily relied on handcrafted spatial and temporal features. 
Subsequently, hybrid models like CNN-TLVQM~\cite{korhonen2020blind} and RAPIQUE~\cite{tu2021rapique} combined these handcrafted features with representations learned by pretrained CNNs. 
More recently, purely learning-based models become more and more popular. 
One common strategy is to employ pretrained networks as separate feature extractors and aggregate features by a subsequent regression module, where relevant works include PVQ~\cite{ying2021patch}, and CoINVQ~\cite{wang2021rich}. 
Further advancements include specialized end-to-end techniques, such as the fragment-based processing in FastVQA~\cite{wu2022fast}, the incorporation of aesthetic features in DOVER~\cite{wu2022disentangling}, and the use of modular designs targeting specific quality factors like content, resolution, and frame rate in ModularBVQA~\cite{wen2024modular}.

\noindent\textbf{MLLM-based Methods.} 
Different from traditional VQA models, MLLM-based methods leverage the extensive knowledge embedded within MLLMs to provide detailed descriptions related to video content and enhance the interpretability of quality assessment. One branch of this research focuses primarily on improving the descriptive capabilities of MLLMs for VQA tasks. For instance, Q-Instruct~\cite{wu2024qinstruct} and DepictQA~\cite{you2024depicting} employ specialized datasets and tailored fine-tuning strategies to enhance the low-level perceptual abilities required for nuanced quality descriptions. Conversely, other approaches prioritize direct quality score prediction. Models like Q-Align~\cite{wu2024qalign} harness the perceptual understanding of MLLMs specifically to generate accurate scores.

\section{CP-LLM: Context- and Pixel-aware Large Language Model}
\subsection{Model Components}
To achieve good context understanding capability and good sensitivity to pixel distortions, the proposed CP-LLM model incorporates an MLLM architecture with dual vision encoders: one encoder is dedicated to extracting high-level semantic information, while the other focuses on capturing low-level, fine-grained texture details. 
The overall structure of CP-LLM is depicted in Fig~\ref{fig:model}, and the details of four major components are discussed in the following subsections.

\subsubsection{High-level Encoder}

We denote a video sequence as $\bm{x}= \{\bm x_i\}_{i=0}^{N-1}$, where $\bm x_i\in\mathbb{R}^{H\times W \times 3}$ represents the $i$-th frame, $H$ and $W$ are the original frame height and width, and $N$ is the total number of frames in the sequence.
% Following a common strategy for computational efficiency, 
We first uniformly sample a sparse set of $M$ key frames, denoted as $\bm y = \{\bm y_i\}_{i=0}^{M-1}$, from the original sequence $\bm{x}$. Each selected key frame $\bm y_i \in \mathbb{R}^{H\times W \times 3}$ undergoes preprocessing. A vision encoder $\bm{f}_h(\cdot)$
% , specifically a SigLIP~\cite{chen2023pali}, 
processes the input frame $\bm y_i$ after bilinear resizing to $H_h \times W_h$. This encoder outputs a sequence of contextualized visual embeddings $\bm u_i = \bm{f}_h(\bm y_i)$. $\bm u_i$ is then projected through a linear layer to align with a target dimension $D$ (specifically, the embedding dimension of the language decoder), resulting in $\bm u_i \in \mathbb{R}^{T_h \times D}$. 
\vspace{-12pt}
\subsubsection{Low-level Encoder}

The frame $\bm y_i$ is resized using bilinear interpolation to a fixed resolution $H_l \times W_l$. Subsequently, we extract $K$ non-overlapping patches, $\mathcal{P}_i = \{\bm{p}_{i,j}\}_{j=0}^{K-1}$, uniformly from the resized frame, where each patch $\bm{p}_{i,j} \in \mathbb{R}^{H_p \times W_p \times 3}$ has a size $H_p \times W_p$.
These $K$ patches for each key frame $\bm{y}_i$ are then fed into a vision Transformer (ViT) model~\cite{dosovitskiy2020image},
% selected from VideoPrism family~\cite{zhao2024videoprism},
denoted as $\bm{f}_l(\cdot)$. $\bm{f}_l$ processes the patch sequence $\mathcal{P}_i$ and outputs a corresponding sequence of feature embeddings $\bm{h}_i = \{\bm{h}_{i,j}\}_{j=0}^{K-1}$, where $\bm{h}_{i,j} \in \mathbb{R}^{T_q \times  D_q}$, $T_q$ is the number of tokens and $D_q$ is the feature dimension of $\bm{f}_l$. 
The features $\bm{h}_{i,j}$ are then projected through a linear layer to align with a target dimension $D$ as well.
To obtain a fixed-size representation for the frame while reducing the sequence length, we apply average pooling across the patch dimension and followed by a layer normalization, resulting in the final embeddings $\bm v_i \in \mathbb{R}^{T_l \times D}$.

\subsubsection{Language Decoder}

To formalize the language decoder process, we denote the input as a pair $(\bm y_i, \bm t_i)$, where $\bm t_i$ represents the $i$-th input text prompt, typically provided as a sequence of tokens $\bm t_i = \{\bm t_{i,j}\}_{j=0}^{L-1}$, where $L$ is the maximum length of the sequence. The proposed quality-aware vision-language model, which we can represent as a function $\bm g(\cdot, \cdot)$, takes the image $\bm y_i$ and the text $\bm t_i$ as input, and generates an output text sequence $\bm z_i = \{z_{i,j}\}_{j=0}^{O-1}$, where $O$ is the maximum length of the generated output.

\subsubsection{Quality Predictor}

Our video quality predictor, denoted by the function $f(\cdot):\mathbb{R}^{(T_h+T_l)\times D}\mapsto \mathbb{R}$, takes the concatenation of $\bm u_i$ and $\bm v_i$ as input and outputs a scalar quality score $q_i$.
The quality predictor is a regression head, implemented as a two-layer multilayer perceptron (MLP) with a ReLU activation between the fully connected layers, takes the visual embeddings as input and predicts a per-frame quality score. The overall quality score $q$ for the video sequence $\bm x$ is computed as the mean of the quality scores obtained from the $M$ key frames.

\subsection{Robust Multi-task Training Pipeline}
The proposed CP-LLM model employs a multi-task training pipeline designed to integrate learning from diverse supervisory signals. 
This pipeline concurrently optimizes model parameters based on three core objectives central to comprehensive quality assessment:

First, a pairwise quality ranking loss, $\mathcal{L}_1$, encourages the model to correctly order the perceived quality between video variants within a pair:
\begin{equation}
\begin{aligned}
\label{eq:loss_rank}
\mathcal{L}_1 = \frac{1}{\mathcal{B}} \sum_{i=1}^{\mathcal{B}} \mathop{\text{max}}(0, -(q_i^{(1)} - q_i^{(2)})(\hat{q}_i^{(1)} - \hat{q}_i^{(2)})) ,
\end{aligned}
\end{equation}
where $\mathcal{B}$ denotes the batch size, $q_i$ represents the ground-truth quality score (either a mean opinion score (MOS) for an original video or a pseudo-MOS for a generated variant), and $\hat{q}_i$ is the predicted score. Superscripts $(1)$ and $(2)$ indicate the first and second video in the $i$-th pair, respectively.

\begin{table}[tb]\small%\footnotesize
\centering
\caption{MOS/DMOS correlations and flip rates evaluated on the YT-Compress dataset. The top-$2$ results are highlighted in \textbf{bold} and \underline{underline}. 
The numbers for MOS and DMOS are presented in the SRCC / PLCC format.
}
\begin{tabular}{lccc}
\toprule
 Model & MOS  & DMOS & FR$\downarrow$\\
% \whline
\midrule
FastVQA~\cite{wu2022fast}  & 0.759 / 0.768 & 0.272 / 0.380 & 0.220 \\
DOVER~\cite{wu2022disentangling} & 0.783 / 0.798 & 0.399 / 0.516 & 0.170 \\
Q-Align~\cite{wu2024qalign} & \underline{0.795} / \underline{0.799} & \underline{0.507} / \underline{0.581}  & \underline{0.118} \\
CP-LLM (ours)   & \textbf{0.801} / \textbf{0.800} & \textbf{0.540} / \textbf{0.605} & \textbf{0.085} \\
\bottomrule
\end{tabular}
\label{tab:ytb_vp9}
\vspace{-9pt}
\end{table}

Second, a direct quality score regression loss, $\mathcal{L}_2$, focuses on accurately predicting the absolute quality score for original videos:
\begin{equation}
\begin{aligned}
\label{eq:loss_mse}
\mathcal{L}_2 = \frac{\sum_{i=1}^{\mathcal{B}} \mathbb{I}_{q_i^{(1)} \in \mathcal{M}} (q_i^{(1)} - \hat{q}_i^{(1)})^2}{\mathop{\text{max}}( \sum_{i=1}^{\mathcal{B}} \mathbb{I}_{q_i^{(1)} \in \mathcal{M}}, 1)} ,
\end{aligned}
\end{equation}
where $\mathcal{M}$ is the set of ground-truth MOS values for original videos. The indicator function $\mathbb{I}_{q_i^{(1)} \in \mathcal{M}}$ ensures $\ell_2$ loss is calculated only for pairs where the first video is an original variant.

Third, an autoregressive text generation loss, $\mathcal{L}_3$, trains the model to produce relevant natural language descriptions of content and quality attributes:
\begin{equation}
\begin{aligned}
\label{eq:loss_llm}
\mathcal{L}_3 = - \frac{1}{\sum_{i=1}^{\mathcal{B}} O_i} \sum_{i=1}^{\mathcal{B}} \sum_{k=1}^{O_i} \log \hat{p}_{i, k}(z_{i, k}),
\end{aligned}
\end{equation}
where $O_i$ is the length of the target text sequence $z_i$ for the $i$-th sample, and $\hat{p}_{i, k}(z_{i, k})$ is the predicted probability of the $k$-th target token. This standard cross-entropy loss for next-token prediction is applied only to the generated token sequence representing the descriptive output.
The total loss function $\mathcal{L}$ is a combination of these individual losses.

\begin{table*}[tb]\small%\footnotesize
\centering
\caption{Cross-dataset testing of our model against four competing models, all retrained on the official training split of the large-scale LSVQ dataset and tested on other VQA datasets without fine-tuning. 
}
\begin{tabular}{lcccc}
\toprule
 SRCC/PLCC$\uparrow$   & YouTube-UGC & LIVE-YT-Gaming & Shorts-SDR & Shorts-HDR2SDR \\
\midrule
FastVQA~\cite{wu2022fast}   & 0.730 / 0.747 & 0.619 / 0.666 & \textbf{0.791} / 0.800 & 0.530 / 0.652 \\
DOVER~\cite{wu2022disentangling} & 0.777 / 0.792 & \textbf{0.668} / \textbf{0.726} & \underline{0.768} / 0.799 & 0.512 / 0.614 \\
ModularBVQA~\cite{wen2024modular}   & 0.788 / 0.804 &  0.600 / 0.696 & 0.753 / \underline{0.802} & 0.559 / \underline{0.686} \\
Q-Align~\cite{wu2024qalign}  & \textbf{0.833} / \textbf{0.846} &  0.611 / 0.687 & 0.746 / 0.779 & \underline{0.585} / 0.672 \\
CP-LLM (ours)    & \underline{0.810} / \underline{0.813} & \underline{0.620} / \underline{0.702} & 0.759 / \textbf{0.804} & \textbf{0.586} / \textbf{0.708} \\
\bottomrule
\end{tabular}
\label{tab:cross}
\end{table*}

\section{Experiments}

\subsection{Experimental Setups}

\noindent\textbf{Datasets}.
The proposed model, CP-LLM, was trained on LSVQ~\cite{ying2021patch} and augmented with compression variants. These variants were generated by encoding the original LSVQ videos using the H.264 codec at $20$ distinct constant rate factor (CRF) levels, ranging from $15$ to $53$. 
To train the language decoder, videos from LSVQ were annotated using Gemini 2.0, prompted with: ``\textit{Based on the provided video frames, create a three-sentence summary. First, describe the visual content of the video in detail. Second, identify the style of the video. Third, assess the technical quality.}'' 
For evaluation, the performance of CP-LLM was assessed on five diverse UGC datasets selected for their varied content types and distortion characteristics: YT-Compress~\cite{wang2019youtube}, YouTube-UGC~\cite{wang2019youtube}, LIVE-YT-Gaming~\cite{yu2022subjective}, Shorts-SDR~\cite{wang2024youtube}, and Shorts-HDR2SDR~\cite{wang2024youtube}.

\begin{table*}[tb]\small%\footnotesize
\centering
\caption{
Flip rate test results for our proposed model versus five competing models, evaluated on LSVQ-test-1080p and its compression variants. ``Answer / Score'' denotes if the model's comparison output is a textual answer or a numerical score; ``Diff'' indicates the CRF difference between compression pairs. 
}
\begin{tabular}{lccccccc}
\toprule
FR$\downarrow$ & Answer / Score &Diff=$2$ & Diff=$4$ & Diff=$6$ & Diff=$8$ & Diff=$10$ & Diff=$20$ \\
\midrule
Gemini 2.0   & Answer & 0.502  & 0.434 & 0.385 & 0.348 & 0.319 & 0.181 \\
CoINVQ~\cite{wang2021rich} & Score & 0.167  & \underline{0.142} & 0.123 & 0.108 & 0.086 & \underline{0.006} \\
FastVQA~\cite{wu2022fast} & Score & 0.256 & 0.205 & 0.169 & 0.150 & 0.111 & 0.024 \\
DOVER~\cite{wu2022disentangling} & Score &  0.268 & 0.218 & 0.179 & 0.156 & 0.126 & 0.026\\
Q-Align~\cite{wu2024qalign} & Score & \underline{0.162}  & \underline{0.142} & \underline{0.122} & \underline{0.105} & \underline{0.082} & 0.008 \\
CP-LLM (ours) & Score & \textbf{0.125}  & \textbf{0.069} & \textbf{0.048} & \textbf{0.033} & \textbf{0.021} & \textbf{0.002} \\
\bottomrule
\end{tabular}
\label{tab:pairwise}
% \vspace{-3pt}
\end{table*}

\noindent\textbf{Evaluated VQA models}.
We compare our proposed model against a representative set of SOTA VQA methods. This includes several prominent deep learning-based models: 
CoINVQ~\cite{wang2021rich}, FastVQA~\cite{wu2022fast}, and DOVER~\cite{wu2022disentangling}. 
To benchmark with recent multimodal approaches, we also include the MLLM-based method Q-Align~\cite{wu2024qalign} and one commercial MLLM Gemini 2.0.

\noindent\textbf{Implementation Details.}
We selected SigLIP~\cite{chen2023pali} as the high-level encoder, VideoPrism~\cite{zhao2024videoprism} as the low-level encoder, and Gemma~\cite{team2024gemma} as the language decoder.
We set the number of key frames $M=5$ per video for LSVQ, Shorts-SDR and Shorts-HDR2SDR,  $M=8$ for LIVE-YT-Gaming, $M=19$ for YT-Compress and YouTube-UGC. Key frames were processed via two pipelines: 1) bilinear resizing to $448 \times 448$ pixels, ignoring aspect ratio for the semantic vision encoder, and 2) bilinear resizing preserving aspect ratio to fit within $540 \times 1080$ pixels, followed by cropping $K=8$ non-overlapping $224 \times 224$ pixel patches for the low-level vision encoder. 
The maximum sequence length was set to $L=512$ tokens. 
We fine-tuned the language decoder using low-rank adaptation (LoRA), applying a rank of $R=4$ to its query, key, and value attention projections. 
Optimization was performed with the multi-task loss with equal weights using the Adam optimizer for $500$ epochs with a batch size of $128$ and a learning rate of $\eta = 1 \times 10^{-4}$ on a $16$-core TPU v5. 

\subsection{Experiment Results}

Table~\ref{tab:ytb_vp9} compares the VQA model accuracy and robustness on the YT-Compress dataset, evaluating by SRCC/PLCC on mean opinion scores (MOS) and differential mean opinion scores (DMOS), and flip rate (FR) on pairs of compression variants. We can see CP-LLM achieves superior performance on all three dimensions. 
Specifically, it surpasses Q-Align, another MLLM-based approach.
This suggests that CP-LLM not only excels in making accurate quality assessments for single videos but is also sensitive to detecting distortions within the same content.
Moreover, although trained exclusively with H.264 compression, the model demonstrates notable robustness to the VP9 compression present in the YT-Compress dataset.

The cross-dataset MOS correlations are shown in Table~\ref{tab:cross}. These four datasets represent highly challenging and recent VQA benchmarks, with content that diverges substantially from the LSVQ training set. CP-LLM achieves a top-$2$ ranking across these four datasets, demonstrating satisfactory generalization capabilities.

Table~\ref{tab:pairwise} shows the flip rate results observed across $20$ distinct compression variants of the LSVQ-test-1080p dataset, aiming to evaluate the fine-grained distortion sensitivity. 
CP-LLM showcases superior sensitivity to pixel-level distortions, outperforming all other models. This highlights the efficacy of CP-LLM's architecture and comprehensive training strategy in capturing pixel-level artifacts.

\section{Conclusion}\label{sec:conclusion}
We have presented CP-LLM, a novel MLLM-based framework for VQA that simultaneously predicts quality scores and generates interpretable textual descriptions. Our method addresses key challenges in VQA, particularly the perception of pixel-level distortions and the integration of quantitative and qualitative feedback, leveraging dual vision encoders and a multi-task training strategy. Comprehensive experiments verify the effectiveness of CP-LLM's design in achieving robust video quality assessment. Additionally, the model's dual-output architecture provides a comprehensive insight into video quality beyond a single score.

\begin{spacing}{0.95}

\bibliographystyle{IEEEbib}
\bibliography{strings,refs}
\end{spacing}

\end{document}